\documentclass[letterpaper, 10 pt, conference]{ieeeconf}  

\IEEEoverridecommandlockouts                              

\usepackage{url}

\usepackage[hidelinks]{hyperref}

\usepackage{graphicx}
\usepackage{comment}
\usepackage{balance}
\usepackage{subcaption}

\usepackage{amsmath}
\usepackage{amssymb}

\usepackage{multirow}

\usepackage{booktabs}
\usepackage{tabularx}
\usepackage{pifont}
\usepackage{xcolor}

\usepackage{cite}

\usepackage[ruled,vlined,linesnumbered]{algorithm2e}

\newcommand{\tabonesize}[2]{%
  {\footnotesize\selectfont #1} %
   ({\scriptsize\selectfont #2})%
}

\newcommand{\tabtwosize}[2]{%
  {\footnotesize\selectfont #1} %
   ({\scriptsize\selectfont #2})%
}

\newcommand{\tabthreesize}[2]{%
  {\footnotesize\selectfont #1} %
   ({\scriptsize\selectfont #2})%
}
\newcommand{\tabfoursize}[2]{%
  {\footnotesize\selectfont #1} %
   ({\scriptsize\selectfont #2})%
}

\title{\LARGE \bf
Mind the Context: Continual Learning of Socially Appropriate Robot Actions via Environmental–Social Disentanglement \vspace{-0.5em}
}

\author{Rafal Robert Karpinski$^{1, *}$, Fethiye Irmak Dogan$^{2, *}$, Nikhil Churamani$^{2}$, Yiming Luo$^{2}$, \\Maartje M.A. de Graaf$^{1}$, 
Davide Dell'Anna$^{1}$, 
Hatice Gunes$^{2}$
\thanks{$*$ Both authors contributed equally to this research.}%
\thanks{$^{1}$ Utrecht University, Utrecht, The Netherlands. {\tt\small{rrk@students.}, \tt\small {d.dellanna@, m.m.a.degraaf@}(uu.nl)}.}%
\thanks{$^{2}$ Department of Computer Science and Technology, University of Cambridge, UK. {\tt\small {fid21, hg410}(@cam.ac.uk)}.}
\thanks{This research was undertaken while R. Karpinski was a remote visiting MSc student at the AFAR Lab, Dept. of CST, University of Cambridge. \textbf{Funding:} The work of F. I. Dogan, N. Churamani, Y. Luo \& H. Gunes were supported by Google via a Google Initiated Grant (GIG). 
\textbf{Open access:} For the purpose of open access, the authors have applied a Creative Commons Attribution (CC BY) license to any Accepted Manuscript version arising. 
\textbf{Contributions:} Conceptualisation, Project administration: FID, HG. Methodology, Formal analysis, Visualisation \& Writing – original draft: RRK, FID. Dataset creation, curation \& labelling: NC, YL, HG. Software \& Investigation: RRK. Writing – review \& editing: RRK, FID, DA, MG, HG. Supervision: FID, DDA, HG. Resources \& Funding acquisition: HG \& NC.}%
}%

\begin{document}

\maketitle
\thispagestyle{empty}
\pagestyle{empty}


\begin{abstract}

Social robots are expected to operate across diverse environments, where similar arrangements can imply different socially appropriate actions, 
e.g., starting a conversation may be acceptable in a crowded home but disruptive in an office meeting. Because such norms and environments cannot all be anticipated in advance, robots require continual learning (CL) to adapt from sequential experience while retaining previously acquired knowledge. Prior work has studied CL for generating socially appropriate robot actions, but it has not addressed domain-incremental settings in which the robot incrementally encounters diverse contexts (e.g., living room, meeting room, office, hallway), where both environmental (e.g., whether the space is open or cluttered with furniture) and social cues (e.g., how people or other agents are positioned around the robot) jointly shape the appropriateness of robot actions. 
We address this gap with the Explicit Disentanglement Dual-Branch (EDD) framework. EDD explicitly separates environmental and social-agent related knowledge and uses replay-based rehearsal to mitigate forgetting while learning the appropriateness of robot actions (e.g., cleaning, serving, starting a conversation) across several indoor domains. 
Experiments show that EDD outperforms several state-of-the-art baselines, and ablation studies further evaluate different disentanglement strategies and the sensitivity to domain ordering. Our code is publicly available at \href{https://github.com/Cambridge-AFAR/Mind-the-Context.git}{https://github.com/Cambridge-AFAR/Mind-the-Context.git}.



\end{abstract}

\section{INTRODUCTION} \label{sec:intro}
\vspace{-0.3em}

\begin{figure}
    \centering
    \includegraphics[width=1\linewidth]{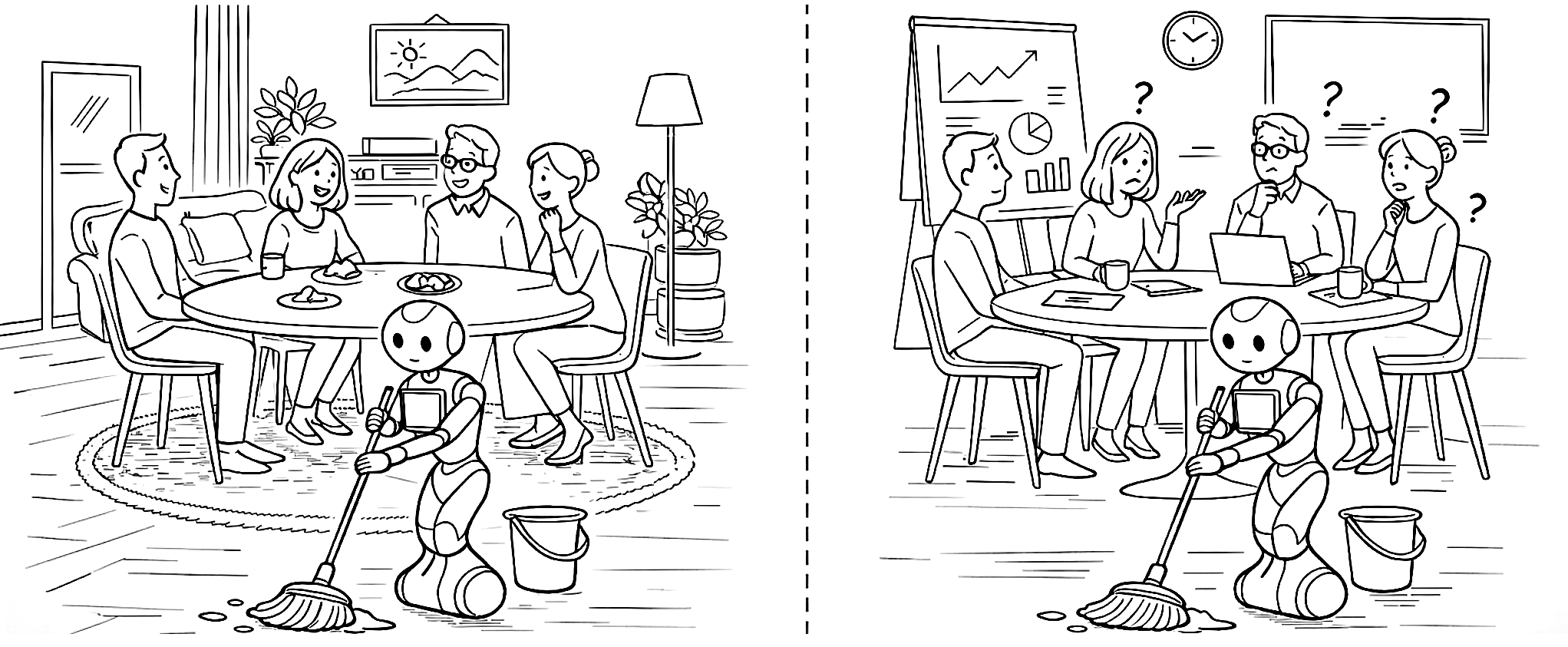}
    \caption{\small Even with similar social arrangements, what is socially appropriate is environment-dependent. Robot cleaning can be appropriate in a living room (left), inappropriate in a meeting (right).}
    \label{motivate}
    \vspace{-1em}
\end{figure}

Social robots must operate across large, diverse indoor spaces, from homes to offices, where similar spatial arrangements can imply different socially appropriate robot actions. For example, a crowded living room may indicate a casual gathering where serving food or starting a conversation is appropriate, while a crowded office room may indicate an ongoing meeting where the same action would be disruptive -- see Figure~\ref{motivate} for an illustrative example. 
Such context-dependent norms are difficult to encode in handcrafted systems~\cite{colledanchise_behavior_2018, iovino_survey_2022}, and it is unrealistic to assume that robots can be trained on all environmental settings and norm variations they will encounter. Instead, they must continually learn from new environments while retaining their former knowledge.


Continual Learning (CL) addresses learning in changing environments while mitigating catastrophic forgetting~\cite{lesort_continual_2020, ven_continual_2024}. It is especially important in robotics, where robots deployed in the real world face data distribution drift as environments, users, and tasks change over time~\cite{shaheen2022continual, zuffer2025advancements}. In social robotics, CL has been applied to learning user routines~\cite{11217699}, contextual understanding~\cite{dogan_deep_2018, 8593633}, and robots' social appropriateness~\cite{tjomsland_mind_2022, churamani_feature_2024}. However, work on learning socially appropriate robot actions remains limited to single-domain and does not address \emph{domain-incremental} CL.


A central challenge in domain-incremental CL for socially appropriate robot actions is that action appropriateness depends on two complementary sources of information: environmental context and social arrangement~\cite{thompson2025social}. Environmental cues (e.g., layout, openness, and furniture) indicate what activities are expected in a space, while social cues (e.g., the positions of people relative to the robot) constrain which actions are appropriate. Prior approaches for learning socially appropriate robot actions typically use full-scene images, which mix these cues implicitly~\cite{churamani_feature_2024}. We argue that making this separation explicit can help domain-incremental CL by reducing spurious correlations and improving knowledge transfer across environments. To investigate this, we pose three interconnected research questions: whether explicit environmental--social disentanglement improves domain-incremental CL for socially appropriate robot actions (\textbf{R1}), how these cues should be represented and separated in visual observations (\textbf{R2}), and motivated by evidence that learning order can strongly influence CL outcomes~\cite{bell_effect_2022, churamani_clifer_2020}, how the order of encountered domains affects a robot’s CL of socially appropriate actions (\textbf{R3}).

\newcommand{\cmark}{\ding{51}} 
\newcommand{\xmark}{\ding{55}} 


\begin{table*}[t]
\centering
\caption{\small Summary of related work and gaps toward \emph{domain-incremental} continual learning of socially appropriate robot actions.}
\label{tab:rw_gaps_check}
\footnotesize
\setlength{\tabcolsep}{3pt}
\renewcommand{\arraystretch}{0.95}
\begin{tabularx}{\textwidth}{p{3.8cm} p{2.2cm} c p{0.8cm} p{1cm} X}
\toprule
\textbf{Focus \& approach} & \textbf{Representative work} &
\textbf{CL?} & \textbf{CL setting} & \textbf{Explicit env/social} &
\textbf{Main limitation for our problem} \\
\midrule

Socially appropriate robot actions (non-CL; supervised/zero-shot)
& \cite{dogan_grace_2024, zhang2023large, rosen2022norm, wachowiak2024llms, checker_federated_2024}
& \xmark & --- & \xmark
& Studies action appropriateness in fixed settings; does not address continual learning under environmental change. \\
\addlinespace[1pt]

Socially appropriate robot actions (CL; e.g., BNN / federated)
& \cite{tjomsland_mind_2022, churamani_feature_2024}
& \cmark & Task-inc. & \xmark
& Mainly single-domain / task-incremental; does not address domain-incremental continual learning across environments. \\
\addlinespace[1pt]

Domain-incremental CL in robotics (other tasks)
& \cite{11217699, dogan_deep_2018, 8593633, podder2025replay, podder2025uncertainty}
& \cmark & Domain-inc. & \xmark
& Addresses domain shifts, but not socially appropriate robot actions (and no explicit env/social separation). \\
\addlinespace[1pt]

Domain adaptation for transfer (domain-invariant learning)
& \cite{ganin2016domain, wulfmeier_incremental_2018, zhang_adversarial_2021}
& \xmark & --- & \xmark
& Seeks domain invariance; for social norms, environment context can be part of the normative signal. \\
\midrule

\textbf{Socially appropriate robot actions (ours: domain inc.)}
& \textbf{EDD (this work)}
& \textbf{\cmark} & \textbf{Domain-inc.} & \textbf{\cmark}
& \textbf{Domain-incremental CL with explicit env/social separation (panoptic segmentation + dual-branch) and replay-based rehearsal.} \\
\bottomrule
\end{tabularx}
\end{table*}

To address these research questions, we propose the \textbf{Explicit Disentanglement Dual-Branch} (EDD) framework for domain-incremental learning of action appropriateness. EDD uses panoptic segmentation to decompose each scene into two inputs: one highlighting the surrounding environment and one highlighting social agents. A dual-branch network processes these inputs and fuses their representations to learn appropriateness scores for multiple robot actions (e.g., mopping, vacuuming, starting a conversation, carrying objects, etc.). To mitigate forgetting across sequential domains (living room and several office spaces, such as small/large office, meeting room, hallway), EDD uses replay-based rehearsal with a memory buffer~\cite{rolnick_experience_2019}.

We evaluate EDD on six indoor domains spanning home and office settings, compare it with several baselines, and test alternative disentanglement strategies and domain orders. Results show that EDD outperforms state-of-the-art approaches and further highlight the impact of the disentanglement strategy and domain order on continual learning outcomes. These findings suggest broader opportunities for disentanglement-based continual robot learning in long-term, real-world deployments, helping maintain socially appropriate robot actions in changing environments.

\vspace{-0.3em}
\section{RELATED WORK}
\vspace{-0.3em}



Continual learning (CL) is critical for real-world robot learning, where robots must operate in dynamic environments that cannot be fully anticipated in advance~\cite{shaheen2022continual, zuffer2025advancements}. In such conditions, CL enables adaptation under data distribution shifts while preserving previously learned knowledge~\cite{lesort_continual_2020, ven_continual_2024}, and it has been applied to several robotics problems that require sustained adaptation during long-term deployment, such as learning user routines through interactions~\cite{11217699} or contextual understanding over time~\cite{dogan_deep_2018, 8593633}. Existing robotics CL work spans both \emph{task-incremental} and \emph{domain-incremental} settings, depending on whether the main shift is in the task~\cite{tjomsland_mind_2022, churamani_feature_2024, 8911341} or in the environment~\cite{11217699, dogan_deep_2018, 8593633, podder2025replay}. 

CL is especially important for socially appropriate robot actions because social norms are context- and environment-dependent~\cite{lawrence2025role}, and robots must adapt their behaviour as environments change over time. Prior work addressed learning socially appropriate actions in several applications, including on socially aware navigation, where robots adapt motion and proxemic behaviour to nearby people and shared spaces~\cite{chen_socially_2017, tsoi_sean_2020, manso_socnav1_2020, karnan_socially_2022, gao_learning_2019, walters_robotic_2007}, as well as approaches targeting interaction-relevant perception and behaviour such as intent recognition~\cite{losey_review_2018}, and engagement dynamics~\cite{salam_fully_2017, michalowski_spatial_2006}. Recent work shows that \emph{task-incremental} CL methods are feasible for generating socially appropriate robot actions~\cite{tjomsland_mind_2022, churamani_feature_2024}. For example, Tjomsland et al.~\cite{tjomsland_mind_2022} used Bayesian neural networks to continually learn the appropriateness of actions within a spatial region or direction, while Churamani et al.~\cite{churamani_feature_2024} studied a federated CL setup for privacy and computational efficiency. However, these studies are mainly limited to single-domain settings and do not address \emph{domain-incremental} continual learning across changing environments.

Classical domain adaptation methods—often developed for computer vision—typically seek domain-invariant representations to enable transfer across domains~\cite{ganin2016domain, wulfmeier_incremental_2018, zhang_adversarial_2021}. However, for social appropriateness, removing domain-specific context may be counterproductive, since the environment can be part of the normative signal, i.e. environmental, situational and social characteristics jointly influence which actions are deemed acceptable~\cite{fiore2013toward}. Accordingly, there remains a need for approaches that support continual, domain-incremental learning of social appropriateness while retaining both environmental and social knowledge -- see Table~\ref{tab:rw_gaps_check} for identified gaps. In this work, we address this need by explicitly separating environmental and social information through panoptic segmentation and dual-branch processing, 
along with replay-based rehearsal to improve knowledge retention~\cite{rolnick_experience_2019}.

\section{METHODOLOGY}
\vspace{-0.3em}

\begin{figure*}
    \centering
    \includegraphics[width=\linewidth]{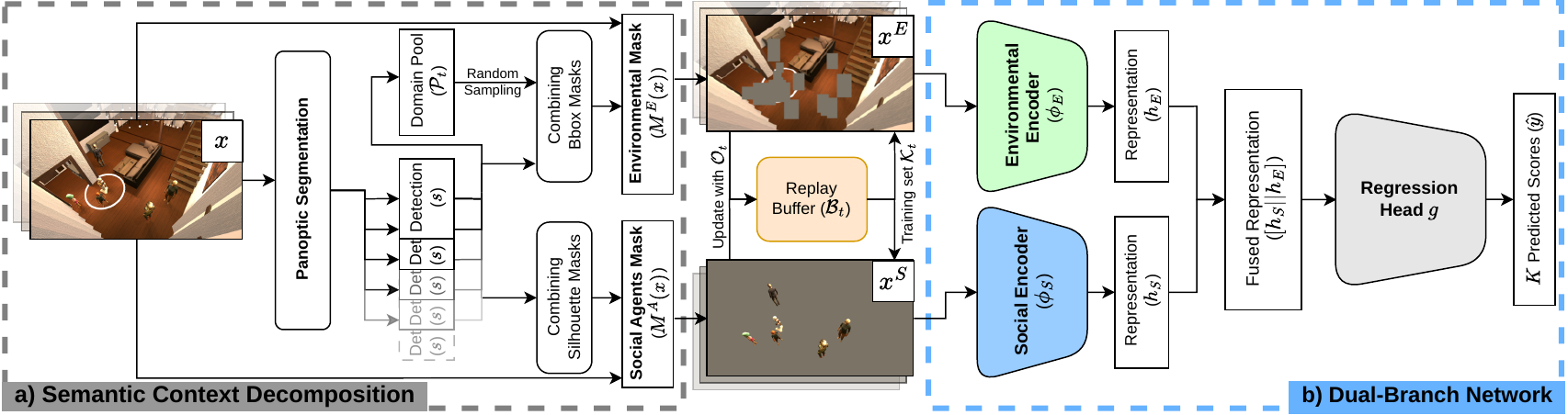}
    \caption{\small Explicit Disentanglement Dual-Branch (EDD) composed of (a) Semantic Context Decomposition and (b) Dual-Branch Network.}
    \label{fig:edd_model}
    \vspace{-2em}
\end{figure*}

\subsection{Problem Formulation}
We address the problem of enabling a robot to continually learn to assess the social appropriateness of its actions 
from visual observations across a sequence of distinct indoor locations. Different from prior work on social appropriateness prediction~\cite{tjomsland_mind_2022, churamani_feature_2024, dogan_grace_2024}, we formulate the problem in the context of domain incremental learning.

Let $\mathcal{A} = \{a_1,\dots,a_K\}$ denote the set of possible robot actions (e.g. starting a conversation, serving food/drinks, carrying small/large objects). For given scene images $\mathcal{X}$, \textcolor{black}{the robot estimates the appropriateness of each action $a_k \in \mathcal{A}$ in the observed scene $x_i \in \mathcal{X}$. Similar to~\cite{tjomsland_mind_2022, churamani_feature_2024, dogan_grace_2024},} we formulate this as a multi-output regression problem in which the robot learns a parametrised function, $f_\theta: \mathcal{X} \rightarrow \mathbb{R}^K$, that maps an input scene $x_i$ to a $K$-dimensional appropriateness score vector: $\hat{\mathbf{y}}_i = f_\theta(x_i) = [\hat{y}_{i,1},\dots,\hat{y}_{i,K}]$, where each component $\hat{y}_{i,k} \in \mathbb{R}$ represents the predicted appropriateness score of action $a_k$ for scene $x_i$. 

We frame the challenge of learning $f_\theta$ over sequentially new locations as a domain-incremental continual learning (DIL) problem. Under this setting, the scenes $\mathcal{X}$ that the robot encounters are partitioned into a sequence $\mathcal{D}_1,\dots,\mathcal{D}_T$, where $T$ denotes the total number of domains and each domain $\mathcal{D}_t \subset \mathcal{X}$ corresponds to a distinct indoor location (e.g., living room, office) exhibiting its own characteristic distribution of scenes and appropriateness labels. Background layout and typical agent configurations differ between locations, and some robot actions are judged appropriate more often in certain locations than in others. 
At time step $t$, the robot operates in location $\mathcal{D}_t$ and has access only to the corresponding training dataset:
\vspace{-0.2em}
\[
\mathcal{O}_t = \{(x_i^{(t)}, \mathbf{y}_i^{(t)})\}_{i=1}^{N_t}, \qquad
x_i^{(t)}\in\mathcal{D}_t,\;\ \mathbf{y}_i^{(t)}\in\mathbb{R}^K,
\vspace{-0.2em}
\]
\noindent where $N_t = |\mathcal{O}_t|$ is the number of annotated scenes from domain $t$ and $\mathbf{y}_i = [y_{i,1},\dots,y_{i,K}]$ are human annotations that are used as ground-truth social appropriateness scores similarly to previous research~\cite{tjomsland_mind_2022, churamani_feature_2024, dogan_grace_2024}. 
As new domains are encountered, the model is updated sequentially, without directly accessing the data from previous domains $\mathcal{O}_{1:t-1}$. Over time, it must accumulate knowledge such that task-appropriateness predictions remain reliable across all encountered domains $\mathcal{O}_{1:t}$.
Conceptually, the ideal training objective is to learn parameters $\theta^\star$ that minimise the cumulative loss across all domains:
\vspace{-0.2em}
\begin{equation}\label{eq:joint_training_objective}
\theta^\star = \arg\min_\theta \sum_{t=1}^{T} \frac{1}{N_t}\sum_{i=1}^{N_t}\ell\big(f_\theta(x_i^{(t)}),\mathbf{y}_i^{(t)}\big),
\vspace{-0.2em}
\end{equation}
\[
\vspace{-0.2em}
 \ell(f_\theta(x),y) = \|f_\theta(x)-y\|_2^2,
\]
\noindent where $\ell(\dots)$ is the per-sample loss and learning $\theta^\star$ corresponds to joint training across all domains if data from all $\mathcal{O}_{1:T}$ were simultaneously available.

\subsection{Domain-incremental learning with rehearsal}
To approximate this ideal objective under the DIL constraint that the robot can only directly access data from the current domain $\mathcal{O}_t$, we adopt a standard Experience Replay strategy~\cite{hsu_reevaluating_2019}. The robot maintains a memory replay buffer $\mathcal{B}_t$ of fixed capacity $B$ that accumulates a subset of training examples from each encountered domain.
When training on domain $\mathcal{D}_t$, we therefore expand the samples available to the robot by mixing the current-domain dataset $\mathcal{O}_t$ with past-domains replay samples from the buffer $\mathcal{B}_{t-1}$. The mixed training set of available knowledge is defined as:
\vspace{-0.2em}
\[
\mathcal{K}_t
= \underbrace{\{(x_i, \mathbf{y}_i)\}_{i=1}^{n}}_{\mathcal{O}_t}
\;\cup\;
\underbrace{\{(x_j, \mathbf{y}_j)\}_{j=1}^{n}}_{\mathcal{B}_{t-1}}.
\vspace{-0.2em}
\]
In our formulation, for $n$ samples from $\mathcal{O}_t$, we draw $n$ replay samples from $\mathcal{B}_{t-1}$. Replay samples are split evenly among all past domains in the buffer, ensuring each contributes the same number of samples.
Finally, we train $f_\theta$ by minimising a mean-squared error (MSE) 
over $\mathcal{K}_t$: 
\vspace{-0.2em}
\[
\mathcal{L}_{t}(\theta)
=
\frac{1}{|\mathcal{K}_{t}|}
\sum_{(x,y)\in \mathcal{K}_{t}}
\ell(f_\theta(x),y).
\vspace{-0.2em}
\]
\noindent where $\mathcal{L}_{t}(\theta)$ represents the loss function. After training on domain $\mathcal{D}_t$, we add its training set to the buffer and then downsample each stored domain to maintain the capacity and equal per-domain quotas; therefore, only a subset of examples from each domain is retained. By training on mixed samples, the model incrementally adapts to the new domain while rehearsing previously seen domains and mitigating catastrophic forgetting.

\subsection{Semantic Context Decomposition} \label{subsec:semantic_context_decomposition}
In addition to DIL, we propose a semantic context decomposition to better capture the complementary environmental and social cues that determine the appropriateness of robot actions. This approach is motivated by the assumption that social reasoning about action appropriateness depends jointly on: (i) the properties of the surrounding environment (affordances and constraints of the setting) - the \textbf{environmental context $x^E$}, and (ii) the configuration of social entities (e.g. humans, robots, pets) in a setting - the \textbf{social context $x^S$}. We formalise this as follows:
\vspace{-0.2em}
\begin{equation} \label{eq:context_decomposition_prior}
p(\mathbf{y}\mid x) \approx p(\mathbf{y}\mid x^E,x^S).
\vspace{-0.2em}
\end{equation}
Accordingly, we encode this structural decomposition at the input level by splitting each scene into environmental- and social-focused views, $x\mapsto (x^E,x^S)$, as demonstrated in Fig.~\ref{fig:edd_model} (a).
To achieve this, we define two deterministic preprocessing transformations $T_E,T_S$ such that:
\vspace{-0.2em}
\[
x^E = T_E(x),\quad x^S= T_S(x).
\vspace{-0.2em}
\]
To obtain $x^E$ and $x^S$, we use a pre-trained \textbf{Panoptic Segmentation} model\footnote{https://github.com/autodistill/autodistill-grounded-sam}.
For each scene $x \in \mathcal{X}$, the model detects a set of social agents $\mathcal{S}(x)$. Specifically, for each detected agent $s \in \mathcal{S}(x)$, it returns two binary masks, $M_s^A, M_s^B \in \{0,1\}^{H \times W}$, representing the agent's silhouette and its bounding box (Bbox), respectively.
Next, for each domain $\mathcal{D}_t$, we construct a shared mask pool defined as
$\mathcal{P}_t=\{\, M_{s,i}^B \mid s \in \mathcal{S}(x_i),\  \ x_i \in \mathcal{O}_t,\}$. Thus, $\mathcal{P}_t$ contains the Bbox masks $M_{s,i}^B$ of all detected agents across all scenes in domain $\mathcal{D}_t$.

To construct the environmental view $x^E$ for scene $x$, we randomly sample $m$ masks from $\mathcal{P}_t$, denoted by $\widetilde{\mathcal{M}}_t^B$, and combine them with the agent Bbox masks $\{M_s^B \mid s \in \mathcal{S}(x)\}$ for that scene. The resulting mask, denoted as the \textbf{Environmental Mask} $M^E(x)$, is designed to (i) hide the regions occupied by the agents in the current scene, (ii) introduce additional sampled Bbox occlusions that further obscure agent locations, and (iii) preserve domain-consistent occlusion patterns. In practice, we apply $M^E(x)$ to the scene by replacing the masked pixels with a fixed RGB value $\mu \in \mathbb{R}^3$:
\[
x^E = T_E(x) = \big(1 - M^E(x)\big)\odot x + M^E(x)\odot \mu
\vspace{-0.2em}
\]
\noindent where $\odot$ denotes pixel-wise multiplication. Thus, $T_E(x)$ produces an image in which social agents are masked out and their locations are occluded, leaving only the unmasked background visible.

On the other hand, to construct the social view $x^S$ for scene $x$, we combine individual agent silhouette masks $\{M_s^A \mid s \in \mathcal{S}(x)\}$ and obtain a single \textbf{Social Agents Mask}, denoted as $M^A(x)$. We then apply $M^A(x)$ to the scene:
\vspace{-0.2em}
\[
x^S = T_S(x) = M^A(x)\odot x + \big(1 - M^A(x)\big)\odot \mu,
\vspace{-0.2em}
\]
Thus, $T_S(x)$ produces an image showing only social agents against a blank, uniform background.


\setlength{\textfloatsep}{6pt plus 1pt minus 2pt}

\subsection{Dual-Branch Network}
\vspace{-0.3em}
To operationalise the proposed semantic context decomposition, we parameterise $f_\theta$ with a dual-branch perception architecture (Fig.~\ref{fig:edd_model} (b)), which processes $x_E$ and $x_S$ separately and then fuses their representations.
Separate processing enables each branch to specialise in environmental or social cues, while the final fusion allows these features to interact before producing appropriateness scores. In our setting, domains correspond to different scene locations (e.g., home, office, etc), and their distinctive cues are primarily present in the environmental view. Accordingly, the dual-branch design allows two branches to differ in their sensitivity to domain shifts. Practically, we embed environmental and social components as follows:
\vspace{-0.2em}
\[
h_E = \phi_E(x^{E}; \theta_E), \quad h_S = \phi_S(x^{S}; \theta_S),
\vspace{-0.2em}
\]
where $\phi_E$ and $\phi_S$ are \textbf{Environmental Encoder} and \textbf{Social Encoder} mapping images to feature vectors $h_E, h_S \in \mathbb{R}^M$. We fuse the learned representations by concatenation and pass them to a \textbf{Regression Head} $g$:
\vspace{-0.2em}
\[
\hat{\mathbf{y}} = f_\theta(x) = g\big([h_S \,\Vert\, h_E]; \theta_g\big),
\vspace{-0.2em}
\]
where $[\cdot \Vert \cdot]$ denotes feature concatenation and $\theta$ represents all the learnable parameters, i.e.,$\{\theta_S, \theta_E, \theta_g\}$.
The head $g$ decodes the fused representation into predicted appropriateness scores $\hat{\mathbf{y}} \in \mathbb{R}^K$ for all actions in $\mathcal{A}$. 

The overall procedure is summarised in Algorithm~\ref {alg:edd}.

\begin{algorithm}[t]
\caption{\small EDD pipeline for a scene $x$ in domain $\mathcal{D}_t$.}
\label{alg:edd}
\footnotesize
\KwIn{Scene image $x$ from domain dataset $\mathcal{O}_t$.}
\KwOut{Appropriateness scores $\hat{\mathbf{y}} = f_\theta(x) \in \mathbb{R}^K$.}

\vspace{0.5em}\tcc{\textbf{\ \ \ \ \ \ \ \  Semantic Context Decomposition}}\vspace{0.2em}
Apply panoptic segmentation to $x$ and detect social agents $\mathcal{S}(x)$\;
Obtain binary masks $M_s^A, M_s^B$ for each $s \in \mathcal{S}(x)$\;
Assign domain pool $\mathcal{P}_t$ as $\{\, M_{s,i}^B \mid s \in \mathcal{S}(x_i),\  \ x_i \in \mathcal{O}_t\}$\;
Sample $m$ bbox masks from $\mathcal{P}_t$, denoted as $\widetilde{\mathcal{M}}_t^B$\;
Construct Env. Mask $M^E(x)$ from $\widetilde{\mathcal{M}}_t^B \cup \{M_s^B \mid s \in \mathcal{S}(x)\}$\;
Construct Social Agents Mask $M^A(x)$ from $\{M_s^A \mid s \in \mathcal{S}(x)\}$\;
Compute
\vspace{-1em}
\[
  x^E = \bigl(1 - M^E(x)\bigr)\odot x + M^E(x)\odot \mu.
\vspace{-1em}
\]
\[
  x^S = M^A(x)\odot x + \bigl(1 - M^A(x)\bigr)\odot \mu,
\vspace{-0.5em}
\]

\vspace{0.5em}\tcc{\textbf{\ \ \ \ \ \ \ \  Dual-Branch Network}}\vspace{0.2em}
Compute environmental and social features:\newline
$h_E = \phi_E(x^E; \theta_E)$\, \quad $h_S = \phi_S(x^S; \theta_S)$\;
Fuse features and predict appropriateness scores:
$\hat{\mathbf{y}} = g([h_S \,\Vert\, h_E]; \theta_g)$\;
\Return $\hat{\mathbf{y}}$
\end{algorithm}

\begin{figure*}[ht!]
    \centering
    \begin{subfigure}{0.20\linewidth}
        \centering
        \includegraphics[width=\linewidth]{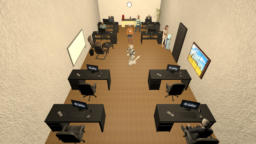}
        \caption{\small Original Image}
        \label{fig:closeup_b}
    \end{subfigure}
        \hfill
    \begin{subfigure}{0.113\linewidth}
        \centering
        \includegraphics[width=\linewidth]{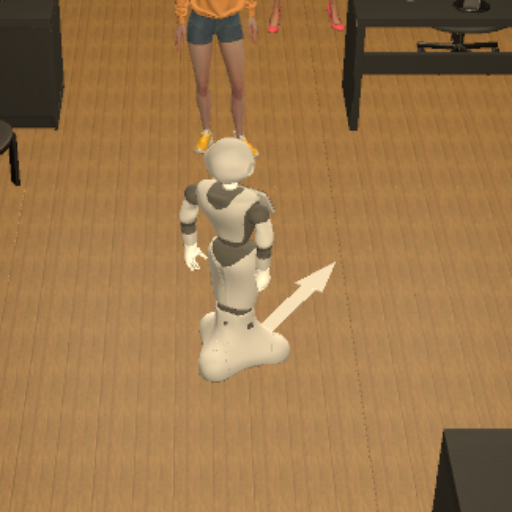}
        \caption{\small  Closed-up}
        \label{fig:closeup_a}
    \end{subfigure}
    \hfill
    \begin{subfigure}{0.20\linewidth}
        \centering
        \includegraphics[width=\linewidth]{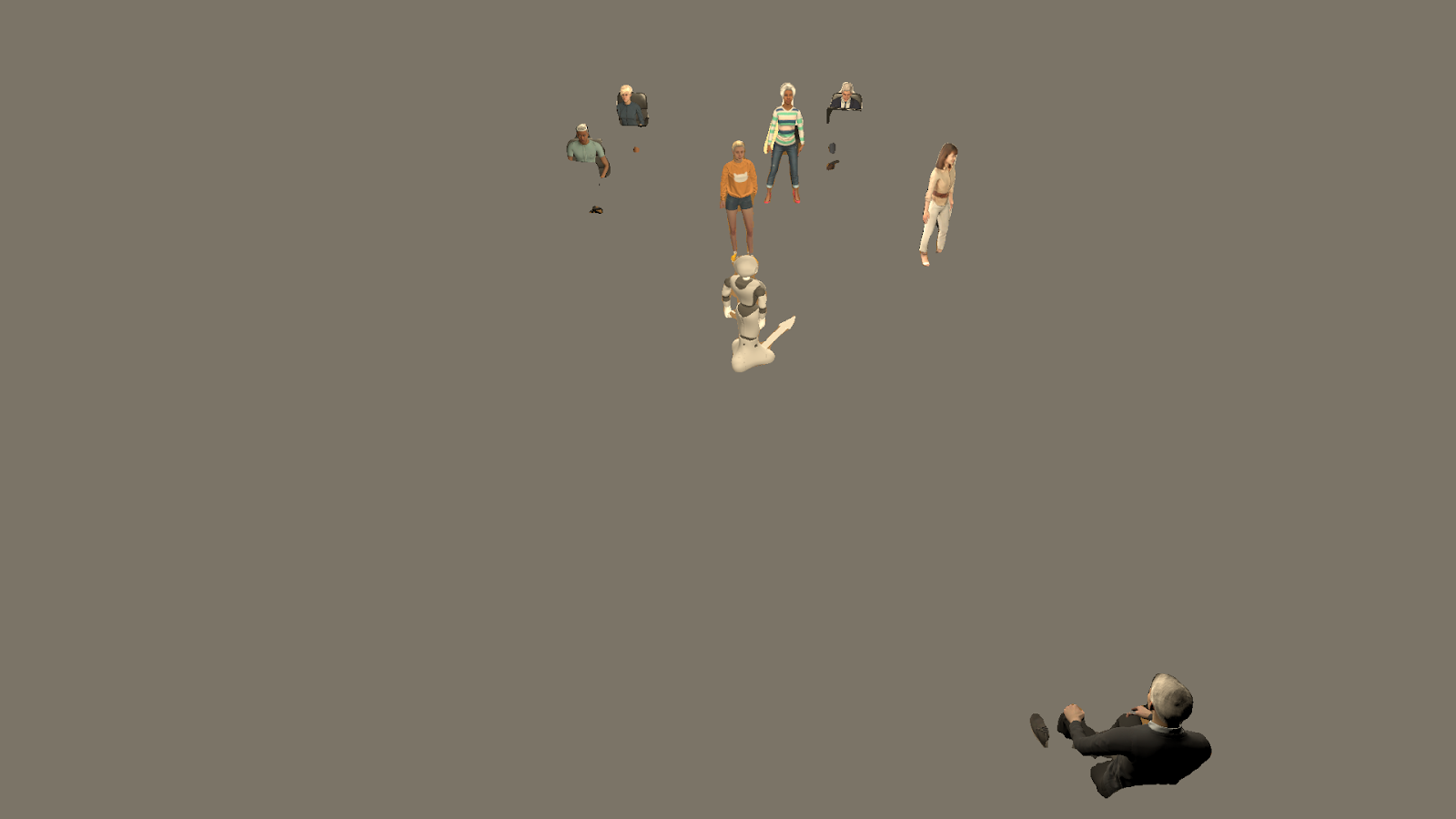}
        \caption{\small  Agents in the scene}
        \label{fig:sil_a}
    \end{subfigure}
    \hfill
    \begin{subfigure}{0.20\linewidth}
        \centering
        \includegraphics[width=\linewidth]{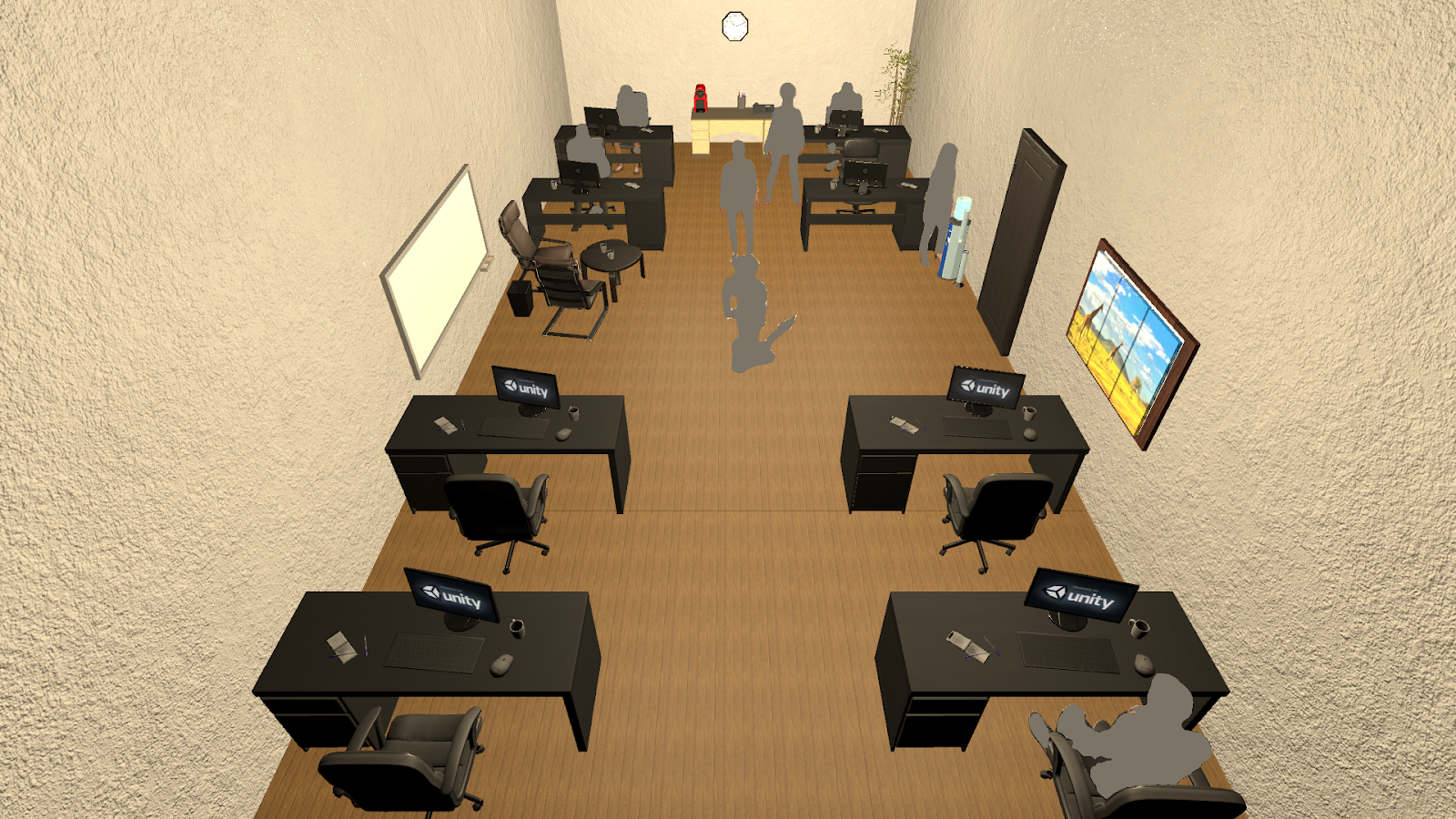}
        \caption{\small Silhouettes}
        \label{fig:sil_b}
    \end{subfigure}
    \hfill
    \begin{subfigure}{0.20\linewidth}
        \centering
        \includegraphics[width=\linewidth]{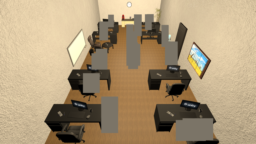}
        \caption{\small Bounding boxes}
        \label{fig:bbox_b}
    \end{subfigure}
    \vspace{-0.5em}
    \caption{\small Images used for different decomposition strategies in dual branches: (i) \textbf{Robot close-up strategy} uses \ref{fig:closeup_b} and \ref{fig:closeup_a}, (ii) \textbf{Silhouette-based decomposition} uses \ref{fig:sil_a} and \ref{fig:sil_b}, (iii) \textbf{Bounding-box decomposition} uses \ref{fig:sil_a} and \ref{fig:bbox_b}, (iv) \textbf{No decomposition control} only uses \ref{fig:closeup_b}.}
    \label{Image_composition}
    \vspace{-0.5em}
\end{figure*}

\section{Experiments}
This section introduces our experiments conducted to investigate research questions introduced in Section~\ref{sec:intro}, along with evaluation datasets, metrics and implementation details. 

\subsection{Dataset}

We evaluate our model on merged OfficeDB and MannersDB+ datasets, which are extensions of MannersDB~\cite{tjomsland_mind_2022} for varying environmental contexts and robot embodiments. Specifically, these datasets depict synthetic images of three service robots (Pepper, NAO, PR2) in varying social configurations and contexts, e.g., robots located around furnished areas with characters representing adults (in group or separate), children and pets, and with music present (see Figure~\ref{fig:edd_model} for home and Figure~\ref{Image_composition} for office scene examples). Prior research used MannersDB+ to learn social appropriateness in a living-room setup~\cite{dogan_grace_2024}.

Overall, these datasets include six distinct indoor environments: \textit{Home} (3000 scenes), and five office-representing spaces, i.e., \textit{BigOffice-2, BigOffice-3, Hallway, MeetingRoom, SmallOffice} (3000 scenes in total, 600 per domain). We use a subset of the dataset with only the Pepper robot to isolate the effects of domain-incremental continual learning without additional confounding factors from multiple robots. This yields 1000 home and 1000 office scenes (200 per office space).

These datasets have been annotated by human raters for social appropriateness of nine robot actions (i.e., vacuum cleaning, mopping the floor, carrying warm food, carrying cold food, carrying drinks, carrying small objects, carrying large objects, cleaning, and starting a conversation) separately, on a scale of 1-5 (one being very inappropriate and five being very appropriate), with three annotations per image on average. 
Similar to Churamani et al.~\cite{churamani_feature_2024}, we mean-aggregated all available annotations per image to obtain a single averaged ground-truth score vector (9-dimensional) representing the social appropriateness of each action. 

\begin{table}[t!]
\centering
\scriptsize
\caption{\small Baselines and rationale for evaluating RQ1. Qwen: Qwen2.5-VL-3B-Instruct, LLaVA: LLaVA-OneVision-0.5B, DeepSeek: DeepSeek-VL-1.3B-Chat.}
\vspace{-0.5em}
\label{tab:rq1_baselines}
\begin{tabular}{p{0.15\linewidth}p{0.06\linewidth}p{0.62\linewidth}}
\hline
\textbf{Model} & \textbf{Type} & \textbf{Rationale} \\
\hline
\textbf{FedLGR}~\cite{churamani_feature_2024}
  & Robotics CL & Task-matched robotics continual learner for socially appropriate actions, extended to our six-domain setup. \\
\hline
\textbf{DUCA}~\cite{gowda2023duca}
  & \multirow{2}{*}{DIL}
  & \multirow{2}{=}{Strong dedicated domain-incremental baselines, adapted to our regression-based setting.} \\
\textbf{DARE++}~\cite{jeeveswaran2024dare} & & \\
\hline
\textbf{Qwen}~\cite{bai2025qwen25vltechnicalreport}
  & \multirow{3}{=}{VLM (zero-shot)}
  & \multirow{3}{=}{State-of-the-art vision–language models used with zero-shot prompts to evaluate their multimodal understanding of socially appropriate robot actions.} \\
\textbf{LLaVA}~\cite{li2024llavaonevisioneasyvisualtask} & & \\
\textbf{DeepSeek}~\cite{lu2024deepseekvl} & & \\
\hline
\textbf{Dual vs }
  & \multirow{2}{=}{EDD variants}
  & \multirow{2}{=}{Full dual-branch model and single-branch ablation, each run under CL, fine-tuning, and joint training to isolate disentanglement and replay effects.} \\
\textbf{Single-branch} & & \\
\hline\hline
\end{tabular}
\end{table}

\vspace{-0.2em}
\subsection{RQ1: Environmental and Social Context Disentanglement}
\vspace{-0.3em}


To evaluate \textbf{RQ1}, the impact of our proposed strategy under domain incremental learning, we compared our Explicit Disentanglement Dual‑Branch Network (EDD) against three types of state-of-the-art baselines: (i) robotics‑specific continual‑learning method for social appropriateness, (ii) recent domain‑incremental computer‑vision methods and (iii) pretrained vision–language models (VLMs) -- see Table~\ref{tab:rq1_baselines} for baseline model details and their rationale.

All baselines were adapted to our domain incremental regression setup. 
For FedLGR~\cite{churamani_feature_2024}, we extended the client-based setup and adjusted output dimensionality. For DUCA~\cite{gowda2023duca} and DARE++~\cite{jeeveswaran2024dare}, we modified 
output heads, reduced batch size and input resolution to satisfy computational constraints. 
\textcolor{black}{All variants shared identical training procedures, replay buffer sizes, and evaluation metrics to ensure comparability.
Motivated by prior zero-shot approach on LLM-based assessment of socially appropriate robot actions~\cite{zhang2023large}, we also queried VLMs with scene images using prompts adapted from that setup.}

\vspace{-0.2em}
\subsection{RQ2: Impact of Disentanglement Strategies}
\vspace{-0.3em}


\begin{table}[t]
\vspace{-0.5em}
\centering
\scriptsize
\caption{\small Input decompositions used to evaluate RQ2.}
\label{tab:rq2_decompositions}
\begin{tabular}{p{0.18\linewidth}p{0.7\linewidth}}
\hline
\textbf{Method} & \textbf{Description} \\
\hline
\textbf{No decomposition control} 
& Single-branch model using the full image with no explicit decomposition (Fig.~\ref{Image_composition} (a)), serving as an experimental control reference for EDD. \\
\hline
\textbf{Robot  \newline close-up } 
& Local–global decomposition uses images (a) and (b) from Fig.~\ref{Image_composition}: crops a robot-centred close-up as a local view, with the full image as a global view, relying on spatially, rather than semantically, defined views. \\
\hline
\textbf{Silhouette \newline decomposition} 
& Uses agents' silhouettes to separate them from the environment. Specifically, it uses images (c) and (d) from Fig.~\ref{Image_composition}, for environmental and social dual branches, respectively. \\
\hline\hline
\textbf{Bounding-box  \newline decomposition \newline (Ours)} 
& Our EDD decomposition uses images (c) and (e) from Fig.~\ref{Image_composition}. It is similar to the Silhouette method, but uses bounding-box masks and additional decoy boxes to make the separation of contexts clearer by obscuring agents’ appearance and positions in the environmental view. \\
\hline
\end{tabular}
\vspace{-0.5em}
\end{table}

To evaluate the impact of different strategies for disentangling context cues (\textbf{RQ2}), we compared four input decomposition strategies: silhouette-based semantic decomposition, bounding-box-based semantic decomposition (ours), spatial decomposition (robot close-up), and a no decomposition control. 
Together, these conditions test (i) whether explicit separation of environmental and social cues improves performance compared to a no decomposition control, (ii) how a semantic decomposition compares to a spatial one, and (iii) whether a clearer semantic separation impacts performance, given that bounding-box masks obscure more of the agents’ appearance and location than silhouettes. 
Figure~\ref{Image_composition} and Table~\ref{tab:rq2_decompositions} summarise experimented disentanglement strategies.

\vspace{-0.2em}
\subsection{RQ3: Sensitivity to the order of domain sequences}
\vspace{-0.3em}

To address \textbf{RQ3}, we examined how the order in which domains are encountered during training affects the performance of EDD.
We therefore considered five domain orders designed to test different intuitions about progression through domains. These include \textit{complexity‑based schedules} that follow a curriculum‑like progression with gradually varying scene complexity, and \textit{alternating‑contrast schedules} that interleave dissimilar domains, which may introduce larger data distribution shifts. While the latter may make learning more challenging, they could also encourage more robust generalisation. The experimental domain orders strategies are summarised in Table~\ref{tab:rq3_orders}. 

\begin{table}[t!]
\centering
\caption{\small Domain orderings used to evaluate RQ3. Home (H), BigOffice-2 (BO2), BigOffice-3 (BO3), Hallway (Hall), MeetingRoom (MR), SmallOffice (SO).}
\scriptsize
\label{tab:rq3_orders}
\begin{tabular}{p{0.14\linewidth}p{0.74\linewidth}}
\hline
\textbf{Order} & \textbf{Description} \\
\hline
\textbf{Default} 
& H→BO2 → BO3 → Hall → MR → SO; OfficeDB default order, starting from spacious home and open offices and ending in more constrained corridor, meeting room, and small office. \\
\hline
\textbf{Complexity$\uparrow$} 
& Hall → SO → MR → BO3 → BO2 → H; progresses from most constrained layouts (Hall, SO) to increasingly spacious and visually complex environments, ending in Home. \\
\hline
\textbf{Complexity$\downarrow$} 
& H → BO2 → BO3 → MR → SO → Hall; reverse of Complexity$\uparrow$, moving from visually rich, open spaces to the most constrained domains. \\
\hline
\textbf{Contrast A} 
& BO3 → SO → H → Hall → MR → BO2; alternating-contrast schedule where adjacent domains are chosen to be highly dissimilar in size, layout, and function. \\
\hline
\textbf{Contrast B} 
& MR → H → Hall → BO2 → SO → BO3; second alternating-contrast schedule following the same high-contrast principle with a different permutation of domains. \\
\hline\hline
\end{tabular}
\vspace{-0.5em}
\end{table}

\subsection{Metrics}

\vspace{-0.3em}

Similar to former work on predicting social appropriateness of robot actions~\cite{tjomsland_mind_2022,churamani_feature_2024,dogan_grace_2024}, we evaluate models using three regression metrics and one continual-learning metric. 
Specifically, prediction error is measured by the \emph{Root Mean Squared Error (RMSE)} between predicted and ground‑truth appropriateness scores (lower is better), macro-averaged across domains. To assess how well models preserve the relative structure of the scores compared to the human evaluators, we report \emph{Pearson’s Correlation Coefficient (PCC)} and \emph{Concordance Correlation Coefficient (CCC)}~\cite{lin_concordance_1989} (higher is better) between 9 predicted and ground-truth scores, macro-averaged across domains. Additionally, to quantify forgetting in the continual‑learning setting, we use \emph{Backwards Transfer (BWT)}~\cite{lesort_continual_2020}, defined as the average change in RMSE on previously learned domains after training on subsequent ones (higher BWT indicates greater forgetting).

\subsection{Implementation Details}
\vspace{-0.3em}

\textbf{Architecture and Initialisation:} Semantic Context Decomposition augments environmental view with $m=10$ randomly sampled Bbox masks (see Section~\ref{subsec:semantic_context_decomposition})\textcolor{black}{, with a per-image latency of $\sim$255 ms.} Dual-Branch Network uses two ImageNet‑pretrained MobileNetV2~\cite{Sandler2018MobileNetV2IR} backbones as environmental and social encoders, followed by global average pooling, flattening, and LayerNorm, producing 1280‑dimensional feature vectors for each branch. The shared decoder is a fully connected regression head that takes the 2560‑dimensional concatenated feature vector and applies 3 linear transformations (in steps 2560, 512, 256) with LayerNorm, ReLU and Dropout(0.3) in between. Then, the final linear layer outputs 9 appropriateness scores. Continual‑learning runs use a naive rehearsal buffer of size 60 samples (5\% of the training set), with capacity allocated approximately equally across all stored domains. 
 \textcolor{black}{The per-image inference latency of the dual-branch design is negligible ($\sim$20 ms vs $\sim$13 ms single-branch).}

\textbf{Training and Evaluation Protocol:} We use 5‑fold cross‑validation. 
We first perform a domain-stratified 80:20 train–test split independently within each domain. 
From the training portion, we further hold out 25\% as a validation set for early stopping. 
All models are trained on the fold‑specific train/validation splits and evaluated on the held‑out test split, and reported metrics are averaged over the five folds. In the continual‑learning setting, domains are presented in the \textit{Default} order. 

For training stability
, we linearly rescale the $[1,5]$ appropriateness scores to the $[0,1]$ range and rescale predictions back to $[1,5]$ for reporting. 
Unless specified otherwise, we resize images from $1080\times1920$ to $144\times256$ \textcolor{black}{during segmentation and decomposition stages} to meet computational constraints. \textcolor{black}{Full-resolution panoptic segmentation accounts for the majority of per-image latency overhead ($\sim$7.2 s).} The aspect ratio is preserved to maintain undistorted spatial relations between social agents, which prior work identified as important for assessing action appropriateness and explicitly modelled via input features~\cite{tjomsland_mind_2022, dogan_grace_2024}. We also normalise image pixel values to the ImageNet mean and standard deviation, as required when using a CNN pretrained on ImageNet\footnote{https://pytorch.org/hub/pytorch\_vision\_mobilenet\_v2/}. 

\textbf{Optimisation and hyperparameters:}
All experiments were implemented in PyTorch and run on a single NVIDIA RTX 2060 GPU with an Intel Core i7‑9750H CPU. We used the Adam optimiser with an initial learning rate of $0.003$, a linear learning‑rate scheduler with 5\% warmup (computed over the total number of optimisation steps per domain), and no weight decay. We applied gradient clipping with a maximum norm of 1.0. Training is performed with mixed precision (FP16 autocast with gradient scaling on CUDA). Unless specified otherwise, we used a batch size of 32 
and trained for up to 60 epochs per domain with early stopping on validation loss (patience 15 epochs, minimum improvement threshold $0.003$). Random seeds were set to 42 for NumPy, PyTorch (CPU and GPU), and data shuffling to support reproducibility. 

\begin{table}[t!]
\centering
\footnotesize
\caption{\small Comparative performance of EDD and baselines. Values are mean (SD in parentheses) over 5-fold cross-validation.}
\setlength{\tabcolsep}{5pt}
\begin{tabular}{l|ccc}
\hline
\textbf{Model} & \textbf{RMSE} $\downarrow$ & \textbf{PCC} $\uparrow$ & \textbf{CCC} $\uparrow$ \\
\hline
FedLGR~\cite{churamani_feature_2024} 
  & \tabonesize{2.091}{0.185}
  & \tabonesize{-0.054}{0.119}
  & \tabonesize{-0.018}{0.037} \\
Qwen2.5-3B-I.~\cite{bai2025qwen25vltechnicalreport} 
  & \tabonesize{0.844}{0.009}
  & \tabonesize{0.216}{0.017}
  & \tabonesize{0.041}{0.005} \\
LLaVA-OV-0.5B~\cite{li2024llavaonevisioneasyvisualtask} 
  & \tabonesize{0.864}{0.007}
  & \tabonesize{-0.158}{0.036}
  & \tabonesize{-0.014}{0.004} \\
DeepSeek-1.3B-C.~\cite{lu2024deepseekvl} 
  & \tabonesize{0.863}{0.011}
  & \tabonesize{0.049}{0.025}
  & \tabonesize{0.005}{0.001} \\
DARE++~\cite{jeeveswaran2024dare} 
  & \tabonesize{1.420}{0.096}
  & \tabonesize{0.389}{0.065}
  & \tabonesize{0.131}{0.031} \\
DUCA~\cite{gowda2023duca}
  & \tabonesize{1.273}{0.254}
  & \tabonesize{0.269}{0.266}
  & \tabonesize{0.069}{0.058} \\
EDD (Ours)
  & \tabonesize{\textbf{0.783}}{0.033}
  & \tabonesize{\textbf{0.552}}{0.045}
  & \tabonesize{\textbf{0.375}}{0.041} \\
\hline
\end{tabular}
\label{tab:results_final}
\end{table}

\begin{table}[t]
\centering
\scriptsize
\caption{\small Ablation results comparing (i) \textbf{single-branch} vs \textbf{dual-branch} and (ii) continual learning (\textbf{CL}) vs fine-tuning (\textbf{FT}) vs non‑continual learning (\textbf{joint training}). Values are mean (SD in parentheses) over 5-fold cross-validation. The best non‑joint scores are in bold; joint training acts as an empirical upper bound.}
\setlength{\tabcolsep}{5pt}
\begin{tabular}{ll|cccc}
\hline
\textbf{Input} & \textbf{Tr.} & \textbf{RMSE} $\downarrow$ & \textbf{PCC} $\uparrow$ & \textbf{CCC} $\uparrow$ & \textbf{BWT} $\downarrow$ \\
\hline
Dual-branch & CL 
  & \tabtwosize{\textbf{0.78}}{0.03}
  & \tabtwosize{\textbf{0.55}}{0.05}
  & \tabtwosize{\textbf{0.38}}{0.04}
  & \tabtwosize{\textbf{0.02}}{0.01} \\
Dual-branch & FT
  & \tabtwosize{0.81}{0.02}
  & \tabtwosize{0.54}{0.03}
  & \tabtwosize{0.34}{0.03}
  & \tabtwosize{0.08}{0.02} \\
Single-branch & CL
  & \tabtwosize{0.79}{0.02}
  & \tabtwosize{0.53}{0.05}
  & \tabtwosize{0.36}{0.05}
  & \tabtwosize{0.04}{0.01} \\
Single-branch & FT
  & \tabtwosize{0.82}{0.02}
  & \tabtwosize{0.50}{0.04}
  & \tabtwosize{0.33}{0.03}
  & \tabtwosize{0.10}{0.02} \\
\hline\hline
Dual-branch & Joint
  & \tabtwosize{0.76}{0.02}
  & \tabtwosize{0.57}{0.04}
  & \tabtwosize{0.39}{0.04}
  & -- \\
Single-branch & Joint 
  & \tabtwosize{0.75}{0.01}
  & \tabtwosize{0.58}{0.03}
  & \tabtwosize{0.38}{0.05}
  & -- \\
\hline
\end{tabular}
\label{tab:results_bounds}
\end{table}

\section{RESULTS}
\vspace{-0.2em}

\begin{table*}[ht!]
\centering
\footnotesize
\setlength{\tabcolsep}{2pt}
\caption{\small Comparison of image decomposition strategies: bounding boxes (Bbox), silhouettes (Silh.), robot close‑ups (Close), and a no decomposition control (NoDec) evaluated overall and separately on the Home domain and the
pooled office domains. Values are mean (SD in parentheses) over 5‑fold cross‑validation.}
\begin{tabular}{l|cccc||cccc|cccc}
\hline
& \multicolumn{4}{c||}{\textbf{Overall}}
& \multicolumn{4}{c|}{\textbf{Home}}
& \multicolumn{4}{c}{\textbf{Office}} \\
\textbf{Seq.}
& \textbf{RMSE} $\downarrow$ & \textbf{PCC} $\uparrow$ & \textbf{CCC} $\uparrow$ & \textbf{BWT} $\downarrow$
& \textbf{RMSE} $\downarrow$ & \textbf{PCC} $\uparrow$ & \textbf{CCC} $\uparrow$ & \textbf{BWT} $\downarrow$
& \textbf{RMSE} $\downarrow$ & \textbf{PCC} $\uparrow$ & \textbf{CCC} $\uparrow$ & \textbf{BWT} $\downarrow$ \\
\hline
\textbf{\footnotesize Bbox}
& \tabthreesize{\textbf{0.78}}{0.03} & \tabthreesize{\textbf{0.55}}{0.05} & \tabthreesize{\textbf{0.38}}{0.04} & \tabthreesize{0.02}{0.01}
& \tabthreesize{\textbf{0.88}}{0.04} & \tabthreesize{\textbf{0.53}}{0.04} & \tabthreesize{\textbf{0.37}}{0.02} & \tabthreesize{\textbf{0.02}}{0.02}
& \tabthreesize{\textbf{0.76}}{0.04} & \tabthreesize{\textbf{0.56}}{0.05} & \tabthreesize{\textbf{0.38}}{0.05} & \tabthreesize{0.02}{0.01} \\
\textbf{\footnotesize Silh.}
& \tabthreesize{0.82}{0.04} & \tabthreesize{0.51}{0.04} & \tabthreesize{0.34}{0.03} & \tabthreesize{0.04}{0.03}
& \tabthreesize{0.93}{0.06} & \tabthreesize{0.49}{0.10} & \tabthreesize{0.32}{0.06} & \tabthreesize{0.04}{0.03}
& \tabthreesize{0.80}{0.04} & \tabthreesize{0.51}{0.02} & \tabthreesize{0.35}{0.02} & \tabthreesize{0.04}{0.03} \\
\textbf{\footnotesize Close}
& \tabthreesize{0.79}{0.02} & \tabthreesize{0.51}{0.03} & \tabthreesize{0.34}{0.03} & \tabthreesize{\textbf{0.01}}{0.02}
& \tabthreesize{0.90}{0.06} & \tabthreesize{0.49}{0.06} & \tabthreesize{0.32}{0.03} & \tabthreesize{\textbf{0.02}}{0.03}
& \tabthreesize{0.77}{0.02} & \tabthreesize{0.52}{0.04} & \tabthreesize{0.34}{0.04} & \tabthreesize{\textbf{0.01}}{0.02} \\
\textbf{\footnotesize NoDec}
& \tabthreesize{0.79}{0.02} & \tabthreesize{0.53}{0.05} & \tabthreesize{0.36}{0.05} & \tabthreesize{0.04}{0.01}
& \tabthreesize{0.89}{0.04} & \tabthreesize{0.48}{0.07} & \tabthreesize{0.33}{0.06} & \tabthreesize{0.05}{0.02}
& \tabthreesize{0.77}{0.02} & \tabthreesize{0.55}{0.05} & \tabthreesize{0.36}{0.04} & \tabthreesize{0.03}{0.02} \\
\hline
\end{tabular}
\label{tab:results_disentanglement_strats_perdomain}
\vspace{-1em}
\end{table*}

\begin{table*}[ht!]
\centering
\footnotesize
\setlength{\tabcolsep}{1.8pt}
\caption{\small Comparison of different domain training sequences, evaluated overall and separately on the Home domain and the pooled office domains. Values are mean (SD in parentheses) over 5-fold cross-validation. Def: Default, C$\uparrow$: complexity increasing, C$\downarrow$: complexity decreasing, ConA: contrast A, ConB: contrast B.}
\begin{tabular}{l|cccc||cccc|cccc}
\hline
\small
& \multicolumn{4}{c||}{\textbf{Overall}}
& \multicolumn{4}{c|}{\textbf{Home}}
& \multicolumn{4}{c}{\textbf{Office}}\\
\textbf{Seq.}
& \textbf{RMSE} $\downarrow$ & \textbf{PCC} $\uparrow$ & \textbf{CCC} $\uparrow$ & \textbf{BWT} $\downarrow$
& \textbf{RMSE} $\downarrow$ & \textbf{PCC} $\uparrow$ & \textbf{CCC} $\uparrow$ & \textbf{BWT} $\downarrow$
& \textbf{RMSE} $\downarrow$ & \textbf{PCC} $\uparrow$ & \textbf{CCC} $\uparrow$ & \textbf{BWT} $\downarrow$ \\
\hline
\textbf{Def.}
& \tabfoursize{\textbf{0.78}}{0.03}
& \tabfoursize{\textbf{0.55}}{0.05}
& \tabfoursize{\textbf{0.38}}{0.04}
& \tabfoursize{\textbf{0.02}}{0.01}
& \tabfoursize{0.88}{0.04}
& \tabfoursize{0.53}{0.04}
& \tabfoursize{0.37}{0.02}
& \tabfoursize{\textbf{0.02}}{0.02}
& \tabfoursize{\textbf{0.76}}{0.04}
& \tabfoursize{\textbf{0.56}}{0.05}
& \tabfoursize{\textbf{0.38}}{0.05}
& \tabfoursize{\textbf{0.02}}{0.01} \\
\textbf{C$\uparrow$}
& \tabfoursize{0.80}{0.03}
& \tabfoursize{0.54}{0.04}
& \tabfoursize{\textbf{0.38}}{0.03}
& \tabfoursize{0.03}{0.02}
& \tabfoursize{\textbf{0.84}}{0.02}
& \tabfoursize{\textbf{0.55}}{0.03}
& \tabfoursize{\textbf{0.38}}{0.04}
& \tabfoursize{0.03}{0.06}
& \tabfoursize{0.79}{0.04}
& \tabfoursize{0.54}{0.05}
& \tabfoursize{\textbf{0.38}}{0.04}
& \tabfoursize{0.03}{0.01} \\
\textbf{C$\downarrow$}
& \tabfoursize{0.79}{0.03}
& \tabfoursize{0.53}{0.03}
& \tabfoursize{0.34}{0.02}
& \tabfoursize{0.03}{0.03}
& \tabfoursize{0.89}{0.05}
& \tabfoursize{0.49}{0.07}
& \tabfoursize{0.35}{0.05}
& \tabfoursize{0.04}{0.07}
& \tabfoursize{0.77}{0.03}
& \tabfoursize{0.53}{0.04}
& \tabfoursize{0.34}{0.02}
& \tabfoursize{0.03}{0.02} \\
\textbf{ConA}
& \tabfoursize{0.80}{0.02}
& \tabfoursize{\textbf{0.55}}{0.04}
& \tabfoursize{0.37}{0.03}
& \tabfoursize{0.03}{0.01}
& \tabfoursize{0.93}{0.07}
& \tabfoursize{0.52}{0.03}
& \tabfoursize{0.34}{0.04}
& \tabfoursize{\textbf{0.02}}{0.03}
& \tabfoursize{0.77}{0.02}
& \tabfoursize{0.55}{0.04}
& \tabfoursize{0.37}{0.03}
& \tabfoursize{0.03}{0.01} \\
\textbf{ConB}
& \tabfoursize{0.80}{0.03}
& \tabfoursize{\textbf{0.55}}{0.04}
& \tabfoursize{0.37}{0.04}
& \tabfoursize{0.04}{0.02}
& \tabfoursize{0.93}{0.08}
& \tabfoursize{0.51}{0.06}
& \tabfoursize{0.34}{0.06}
& \tabfoursize{0.03}{0.05}
& \tabfoursize{0.78}{0.03}
& \tabfoursize{0.55}{0.04}
& \tabfoursize{0.37}{0.04}
& \tabfoursize{0.04}{0.03} \\
\hline
\end{tabular}
\label{tab:results_scheduling_domains}
\vspace{-1em}
\end{table*}

\subsection{RQ1: Environmental and Social Context Disentanglement}
\vspace{-0.2em}

Table~\ref{tab:results_final} reports the overall appropriateness prediction performance of EDD (Ours) compared to baseline methods. 
\textcolor{black}{EDD achieves the lowest RMSE and the highest PCC and CCC, indicating more accurate and more consistent action appropriateness predictions than the compared baselines.} 
FedLGR yields the highest error and near-zero PCC and CCC, indicating the weakest overall performance among the compared models. DARE++ achieves the strongest baseline correlations, followed by DUCA. VLM-based methods (Qwen2.5-3B-I., LLaVA-OV-0.5B, DeepSeek-1.3B-C.) yield lower PCC and CCC values.

We further compare EDD ablations across different training regimes (Table~\ref{tab:results_bounds}). Dual‑branch CL (our approach) attains lower error, higher PCC, CCC and lower forgetting (BWT) in comparison to the single-branch ablation that does not use image decomposition. Fine‑tuning without replay degrades performance for both architectures, increasing RMSE, lowering PCC/CCC, and yielding higher BWT. Joint‑training variants, which have access to all domains simultaneously, act as an empirical upper bound, and achieve the best overall accuracy, with the dual‑branch CL model closer to this upper bound than the single‑branch CL model.


\textcolor{black}{Overall, compared to baselines and the non-disentangled variant, EDD shows consistently favourable performance across all metrics, better alignment with human-provided appropriateness scores, and lower forgetting, suggesting improved domain-incremental learning of socially appropriate robot actions.}

\subsection{RQ2: Impact of Disentanglement Strategies}
\vspace{-0.2em}
Table~\ref{tab:results_disentanglement_strats_perdomain} compares alternative image decomposition strategies for the dual-branch network.
Across all splits, the bounding box strategy (Bbox) yields the lowest RMSE and highest PCC/CCC. Robot close-up (Close) and no decomposition (NoDec) obtain similar RMSE, both lower than the Silhouette method (Silh.), while NoDec shows consistently higher correlations than Close and Silh. The forgetting measure (BWT) remains low across all variants, with Bbox and Close achieving the lowest BWT, indicating lower forgetting across domains compared to Silh. and NoDec. Overall, all methods perform relatively similarly, with Bbox achieving the lowest prediction error and strongest correlations.


\subsection{RQ3: Sensitivity to the order of domain sequences}

Table~\ref{tab:results_scheduling_domains} compares EDD under different domain training sequences. Overall, all schedules yield close performance, with RMSE and PCC/CCC in similar ranges, and with decreasing complexity (C$\downarrow$) obtaining the lowest correlations. On Home, the increasing‑complexity schedule (C$\uparrow$) attains the lowest RMSE and highest PCC/CCC, while Default is second‑best and C$\downarrow$/ConA/ConB are noticeably worse. Across the pooled Office domains, Default and the contrastive schedules (ConA/ConB) form a tight top group in RMSE and CCC, with the complexity‑based schedules (C$\uparrow$/C$\downarrow$) yielding slightly higher RMSE and lower PCC. Forgetting (BWT) remains low across all domain orders, with very minor variability (around $\pm 0.01$) between schedules, indicating limited effect on catastrophic forgetting. Overall, domain order has a measurable but modest effect compared to the model architecture or the decomposition strategy.


\section{Discussion, Conclusion and Future Work}

Our work highlights a key insight for continual learning in socially appropriate robotics: making environmental and social cues explicit can support learning action appropriateness under domain shift. In our setting, we address this through the Explicit Disentanglement Dual-Branch (EDD) framework, which separates these cues at the input level and fuses them with a dual-branch architecture trained using replay-based rehearsal. Our results show that the EDD framework outperforms several state-of-the-art baselines and generates appropriateness scores that more closely align with human-provided ones across complementary metrics. This explicit separation also enables targeted analysis of how disentanglement strategies and domain order shape CL performance for socially appropriate actions.

Regarding RQ1, our results show that explicit environmental–social separation supports CL outcomes. Across six domains, EDD achieves the strongest overall performance (Table~\ref{tab:results_final}) and Table~\ref{tab:results_bounds} further demonstrates that the dual-branch with replay memory remains close to the joint-training upper bound. Notably, compared to EDD, VLMs achieve a moderately competitive RMSE but exhibit weak correlation with human-provided scores (PCC/CCC). From a broader robotics perspective, this aligns with prior findings that foundation-model priors alone are insufficient to capture social appropriateness in embodied settings~\cite{dogan_grace_2024}: because both environmental cues (e.g., room layout and affordances) and social cues (e.g., interpersonal spacing and group configuration) jointly determine what is appropriate~\cite{thompson2025social}, structuring perception to retain these cues can be beneficial for learning appropriate robot actions across changing domains.  

The RQ2 findings show that the different decompositions yield broadly similar overall performance, with an important trend: cleaner separation of environmental and social context provides a better match to human-provided judgments (PCC/CCC). 
\textcolor{black}{In this respect, Bbox shows the strongest overall trend (Table~\ref{tab:results_disentanglement_strats_perdomain}), as it more cleanly removes social agents from the environmental context than silhouette masks. }
By contrast, the robot close-up yields the best retention (BWT), likely because its robot-centred view reduces cross-domain interference, while the lower PCC/CCC may reflect that broader social cues, such as grouping and interpersonal spacing, are less visible. Finally, compared to the no decomposition control, the Bbox mask yields better performance, supporting our assumption that decomposing the scene does not discard task-relevant information. 
\textcolor{black}{Overall, modest but consistent gains of Bbox suggest representations that retain socially meaningful context can help CL of socially appropriate robot actions, and losing environmental or social cues can undermine robots' ability to learn human preferences.}


Regarding RQ3, unlike the stronger order effects reported in other CL settings~\cite{bell_effect_2022, churamani_clifer_2020}, domain order has a modest effect in our results (Table~\ref{tab:results_scheduling_domains}). 
\textcolor{black}{The most visible trend appears in Home: the increasing-complexity schedule (C$\uparrow$), which presents Home last, yields the strongest Home performance, suggesting a possible curriculum-style effect~\cite{10.1145/1553374.1553380}, although the small numerical differences mean this should be interpreted as an indicative pattern rather than strong evidence.} 
Effects on the office domains are more mixed, likely because they aggregate several related but distinct spaces. Still, decreasing complexity (C$\downarrow$) performs worst for office environments, specifically for PCC and CCC. 
\textcolor{black}{Thus, while EDD appears relatively robust to ordering in these experiments, domain sequencing remains a plausible factor for future investigation, and order-aware replay strategies could be further explored for continual robot learning of action appropriateness.}



Our work opens several directions for future research.  
For example, EDD can be extended beyond the current single-robot embodiment, and its robustness can be examined under embodiment shifts. 
\textcolor{black}{Future work can further examine EDD in real-world settings to assess its generalisability beyond synthetic scenes, including robustness to synthetic-to-real shifts and the effects of imperfect panoptic segmentation on the environmental--social decomposition.} 
In addition, EDD’s structure naturally invites work on explainability: future work can analyse what each branch learns and how predictions depend on environmental versus social cues. Another promising direction is adaptive fusion, in which the model can dynamically adjust to the separate branches based on task context (e.g., robot action) and interaction environment (e.g., home vs office).

\bibliographystyle{IEEEtran}
\bibliography{references}

@inproceedings{11217699,
 author = {Bartoli, Ermanno and Doğan, Fethiye Irmak and Leite, Iolanda},
 booktitle = {2025 34th IEEE International Conference on Robot and Human Interactive Communication (RO-MAN)},
 doi = {10.1109/RO-MAN63969.2025.11217699},
 number = {},
 pages = {1550-1557},
 title = {STREAK: Streaming Network for Continual Learning of Object Relocations under Household Context Drifts},
 volume = {},
 year = {2025}
}

@inproceedings{8593633,
 author = {Irmak Doğan, Fethiye and Bozcan, Ilker and Celik, Mehmet and Kalkan, Sinan},
 booktitle = {2018 IEEE/RSJ International Conference on Intelligent Robots and Systems (IROS)},
 doi = {10.1109/IROS.2018.8593633},
 number = {},
 pages = {4641-4646},
 title = {CINet: A Learning Based Approach to Incremental Context Modeling in Robots},
 volume = {},
 year = {2018}
}

@article{8911341,
 author = {Feng, Fan and Chan, Rosa H. M. and Shi, Xuesong and Zhang, Yimin and She, Qi},
 doi = {10.1109/ACCESS.2019.2955480},
 journal = {IEEE Access},
 number = {},
 pages = {3434-3441},
 title = {Challenges in Task Incremental Learning for Assistive Robotics},
 volume = {8},
 year = {2020}
}

@article{bai2025qwen25vltechnicalreport,
 author = {Bai, Shuai and Chen, Keqin and Liu, Xuejing and Wang, Jialin and Ge, Wenbin and Song, Sibo and Dang, Kai and Wang, Peng and Wang, Shijie and Tang, Jun and Zhong, Humen and Zhu, Yuanzhi and Yang, Mingkun and Li, Zhaohai and Wan, Jianqiang and Wang, Pengfei and Ding, Wei and Fu, Zheren and Xu, Yiheng and Ye, Jiabo and Zhang, Xi and Xie, Tianbao and Cheng, Zesen and Zhang, Hang and Yang, Zhibo and Xu, Haiyang and Lin, Junyang},
 journal = {arXiv preprint arXiv:2502.13923},
 title = {Qwen2.5-VL Technical Report},
 year = {2025}
}

@article{bell_effect_2022,
  title={The effect of task ordering in continual learning},
  author={Bell, Samuel J and Lawrence, Neil D},
  journal={arXiv preprint arXiv:2205.13323},
  year={2022}
}

@inproceedings{chen_socially_2017,
 author = {Chen, Yu Fan and Everett, Michael and Liu, Miao and How, Jonathan P.},
 booktitle = {2017 {{IEEE}}/{{RSJ International Conference}} on {{Intelligent Robots}} and {{Systems}} ({{IROS}})},
 doi = {10.1109/IROS.2017.8202312},
 month = {September},
 pages = {1343--1350},
 title = {Socially Aware Motion Planning with Deep Reinforcement Learning},
 urldate = {2025-04-03},
 year = {2017}
}

@inproceedings{churamani_clifer_2020,
 author = {Churamani, Nikhil and Gunes, Hatice},
 booktitle = {2020 15th {{IEEE International Conference}} on {{Automatic Face}} and {{Gesture Recognition}} ({{FG}} 2020)},
 copyright = {https://ieeexplore.ieee.org/Xplorehelp/downloads/license-information/IEEE.html},
 doi = {10.1109/FG47880.2020.00110},
 month = {November},
 pages = {322--328},
 title = {{{CLIFER}}: {{Continual Learning}} with {{Imagination}} for {{Facial Expression Recognition}}},
 urldate = {2025-12-10},
 year = {2020}
}

@article{churamani_feature_2024,
 author = {Churamani, Nikhil and Checker, Saksham and Dogan, Fethiye Irmak and Chiang, Hao-Tien Lewis and Gunes, Hatice},
 journal = {arXiv preprint arXiv:2405.15773},
 title = {Feature aggregation with latent generative replay for federated continual learning of socially appropriate robot behaviours},
 year = {2024}
}

@book{colledanchise_behavior_2018,
 address = {USA},
 author = {Colledanchise, Michele and Ogren, Petter},
 edition = {1st},
 isbn = {978-1-138-59373-2},
 month = {June},
 publisher = {CRC Press, Inc.},
 shorttitle = {Behavior {{Trees}} in {{Robotics}} and {{Al}}},
 title = {Behavior {{Trees}} in {{Robotics}} and {{Al}}: {{An Introduction}}},
 year = {2018}
}

@inproceedings{dogan_deep_2018,
 author = {Do{\u g}an, Fethiye Irmak and {\c C}elikkanat, Hande and Kalkan, Sinan},
 booktitle = {2018 {{IEEE International Conference}} on {{Robotics}} and {{Automation}} ({{ICRA}})},
 doi = {10.1109/ICRA.2018.8462925},
 issn = {2577-087X},
 month = {May},
 pages = {2411--2416},
 title = {A {{Deep Incremental Boltzmann Machine}} for {{Modeling Context}} in {{Robots}}},
 urldate = {2025-04-03},
 year = {2018}
}

@inproceedings{dogan_grace_2024,
 author = {Doğan, Fethiye Irmak and Ozyurt, Umut and Cinar, Gizem and Gunes, Hatice},
 booktitle = {2025 IEEE International Conference on Robotics and Automation (ICRA)},
 doi = {10.1109/ICRA55743.2025.11127826},
 number = {},
 pages = {4330-4336},
 title = {GRACE: Generating Socially Appropriate Robot Actions Leveraging LLMs and Human Explanations},
 volume = {},
 year = {2025}
}

@article{fiore2013toward,
 author = {Fiore, Stephen M and Wiltshire, Travis J and Lobato, Emilio JC and Jentsch, Florian G and Huang, Wesley H and Axelrod, Benjamin},
 journal = {Frontiers in psychology},
 pages = {859},
 publisher = {Frontiers Media SA},
 title = {Toward understanding social cues and signals in human--robot interaction: effects of robot gaze and proxemic behavior},
 volume = {4},
 year = {2013}
}

@article{ganin2016domain,
 author = {Ganin, Yaroslav and Ustinova, Evgeniya and Ajakan, Hana and Germain, Pascal and Larochelle, Hugo and Laviolette, Fran{\c{c}}ois and March, Mario and Lempitsky, Victor},
 journal = {Journal of machine learning research},
 number = {59},
 pages = {1--35},
 title = {Domain-adversarial training of neural networks},
 volume = {17},
 year = {2016}
}

@inproceedings{gao_learning_2019,
 author = {Gao, Yuan and Yang, Fangkai and Frisk, Martin and Hemandez, Daniel and Peters, Christopher and Castellano, Ginevra},
 booktitle = {2019 28th {{IEEE International Conference}} on {{Robot}} and {{Human Interactive Communication}} ({{RO-MAN}})},
 doi = {10.1109/RO-MAN46459.2019.8956444},
 issn = {1944-9437},
 month = {October},
 pages = {1--8},
 title = {Learning {{Socially Appropriate Robot Approaching Behavior Toward Groups}} Using {{Deep Reinforcement Learning}}},
 urldate = {2025-04-03},
 year = {2019}
}

@article{gowda2023duca,
 author = {Shruthi Gowda and Bahram Zonooz and Elahe Arani},
 issn = {2835-8856},
 journal = {Transactions on Machine Learning Research},
 title = {Dual Cognitive Architecture: Incorporating Biases and Multi-Memory Systems for Lifelong Learning},
 year = {2023}
}

@article{hsu_reevaluating_2019,
  title={Re-evaluating continual learning scenarios: A categorization and case for strong baselines},
  author={Hsu, Yen-Chang and Liu, Yen-Cheng and Ramasamy, Anita and Kira, Zsolt},
  journal={arXiv preprint arXiv:1810.12488},
  year={2018}
}

@article{iovino_survey_2022,
 author = {Iovino, Matteo and Scukins, Edvards and Styrud, Jonathan and {\"O}gren, Petter and Smith, Christian},
 doi = {10.1016/j.robot.2022.104096},
 issn = {0921-8890},
 journal = {Robotics and Autonomous Systems},
 month = {August},
 pages = {104096},
 title = {A Survey of {{Behavior Trees}} in Robotics and {{AI}}},
 urldate = {2025-12-10},
 volume = {154},
 year = {2022}
}

@inproceedings{jeeveswaran2024dare,
author = {Jeeveswaran, Kishaan and Arani, Elahe and Zonooz, Bahram},
title = {Gradual divergence for seamless adaptation: a novel domain incremental learning method},
year = {2024},
publisher = {JMLR.org},
booktitle = {Proceedings of the 41st International Conference on Machine Learning},
articleno = {862},
numpages = {16},
location = {Vienna, Austria},
series = {ICML'24}
}

@article{karnan_socially_2022,
 author = {Karnan, Haresh and Nair, Anirudh and Xiao, Xuesu and Warnell, Garrett and Pirk, S{\"o}ren and Toshev, Alexander and Hart, Justin and Biswas, Joydeep and Stone, Peter},
 doi = {10.1109/LRA.2022.3184025},
 issn = {2377-3766},
 journal = {IEEE Robotics and Automation Letters},
 month = {October},
 number = {4},
 pages = {11807--11814},
 shorttitle = {Socially {{CompliAnt Navigation Dataset}} ({{SCAND}})},
 title = {Socially {{CompliAnt Navigation Dataset}} ({{SCAND}}): {{A Large-Scale Dataset}} of {{Demonstrations}} for {{Social Navigation}}},
 urldate = {2025-04-01},
 volume = {7},
 year = {2022}
}

@article{lawrence2025role,
 author = {Lawrence, Steven and Jouaiti, Melanie and Hoey, Jesse and Nehaniv, Chrystopher L and Dautenhahn, Kerstin},
 journal = {ACM Transactions on Human-Robot Interaction},
 number = {3},
 pages = {1--44},
 publisher = {ACM New York, NY},
 title = {The Role of Social Norms in Human--Robot Interaction: A Systematic Review},
 volume = {14},
 year = {2025}
}

@article{lesort_continual_2020,
 author = {Lesort, Timoth{\'e}e and Lomonaco, Vincenzo and Stoian, Andrei and Maltoni, Davide and Filliat, David and {D{\'i}az-Rodr{\'i}guez}, Natalia},
 doi = {10.1016/j.inffus.2019.12.004},
 issn = {1566-2535},
 journal = {Information Fusion},
 month = {June},
 pages = {52--68},
 shorttitle = {Continual Learning for Robotics},
 title = {Continual Learning for Robotics: {{Definition}}, Framework, Learning Strategies, Opportunities and Challenges},
 urldate = {2025-01-30},
 volume = {58},
 year = {2020}
}

@misc{li2024llavaonevisioneasyvisualtask,
 author = {Li, Bo and Zhang, Yuanhan and Guo, Dong and Zhang, Renrui and Li, Feng and Zhang, Hao and Zhang, Kaichen and Li, Yanwei and Liu, Ziwei and Li, Chunyuan},
 title = {LLaVA-OneVision: Easy Visual Task Transfer},
 journal = {arXiv preprint arXiv:2408.03326},
 year = {2024}
}

@article{lin_concordance_1989,
 author = {Lin, Lawrence I-Kuei},
 doi = {10.2307/2532051},
 eprint = {2532051},
 eprinttype = {jstor},
 issn = {0006-341X},
 journal = {Biometrics},
 number = {1},
 pages = {255--268},
 publisher = {[Wiley, International Biometric Society]},
 title = {A {{Concordance Correlation Coefficient}} to {{Evaluate Reproducibility}}},
 urldate = {2025-09-04},
 volume = {45},
 year = {1989}
}

@article{losey_review_2018,
 author = {Losey, Dylan P. and McDonald, Craig G. and Battaglia, Edoardo and O'Malley, Marcia K.},
 doi = {10.1115/1.4039145},
 issn = {0003-6900},
 journal = {Applied Mechanics Reviews},
 month = {February},
 number = {010804},
 title = {A {{Review}} of {{Intent Detection}}, {{Arbitration}}, and {{Communication Aspects}} of {{Shared Control}} for {{Physical Human}}--{{Robot Interaction}}},
 urldate = {2025-04-03},
 volume = {70},
 year = {2018}
}

@misc{lu2024deepseekvl,
 archiveprefix = {arXiv},
 author = {Haoyu Lu and Wen Liu and Bo Zhang and Bingxuan Wang and Kai Dong and Bo Liu and Jingxiang Sun and Tongzheng Ren and Zhuoshu Li and Yaofeng Sun and Chengqi Deng and Hanwei Xu and Zhenda Xie and Chong Ruan},
 eprint = {2403.05525},
 primaryclass = {cs.AI},
 title = {DeepSeek-VL: Towards Real-World Vision-Language Understanding},
 year = {2024}
}

@article{manso_socnav1_2020,
 author = {Manso, Luis J. and Nu{\~n}ez, Pedro and Calderita, Luis V. and Faria, Diego R. and Bachiller, Pilar},
 copyright = {http://creativecommons.org/licenses/by/3.0/},
 doi = {10.3390/data5010007},
 issn = {2306-5729},
 journal = {Data},
 langid = {english},
 month = {March},
 number = {1},
 pages = {7},
 publisher = {Multidisciplinary Digital Publishing Institute},
 shorttitle = {{{SocNav1}}},
 title = {{{SocNav1}}: {{A Dataset}} to {{Benchmark}} and {{Learn Social Navigation Conventions}}},
 urldate = {2025-04-01},
 volume = {5},
 year = {2020}
}

@inproceedings{michalowski_spatial_2006,
 author = {Michalowski, M.P. and Sabanovic, S. and Simmons, R.},
 booktitle = {9th {{IEEE International Workshop}} on {{Advanced Motion Control}}, 2006.},
 doi = {10.1109/AMC.2006.1631755},
 issn = {1943-6580},
 month = {March},
 pages = {762--767},
 title = {A Spatial Model of Engagement for a Social Robot},
 urldate = {2025-04-03},
 year = {2006}
}

@article{podder2025replay,
 author = {Podder, Kanchon Kanti and Dutta, Pritom and Zhang, Jian},
 journal = {Electronics},
 number = {19},
 pages = {3946},
 publisher = {MDPI},
 title = {Replay-Based Domain Incremental Learning for Cross-User Gesture Recognition in Robot Task Allocation},
 volume = {14},
 year = {2025}
}

@inproceedings{podder2025uncertainty,
 author = {Podder, Kanchon Kanti and Zhang, Jian and Mao, Shiwen and Yu, Zhitao},
 booktitle = {2025 IEEE 11th World Forum on Internet of Things (WF-IoT)},
 organization = {IEEE},
 pages = {1--7},
 title = {Uncertainty-Aware Domain Incremental Learning for Gesture Recognition in Robotic IoT Systems},
 year = {2025}
}

@article{rolnick_experience_2019,
 author = {Rolnick, David and Ahuja, Arun and Schwarz, Jonathan and Lillicrap, Timothy and Wayne, Gregory},
 journal = {Advances in {{Neural Information Processing Systems}}},
 title = {Experience {{Replay}} for {{Continual Learning}}},
 volume = {32},
 year = {2019}
}

@article{salam_fully_2017,
 author = {Salam, Hanan and {\c C}eliktutan, Oya and Hupont, Isabelle and Gunes, Hatice and Chetouani, Mohamed},
 doi = {10.1109/ACCESS.2016.2614525},
 issn = {2169-3536},
 journal = {IEEE Access},
 pages = {705--721},
 title = {Fully {{Automatic Analysis}} of {{Engagement}} and {{Its Relationship}} to {{Personality}} in {{Human-Robot Interactions}}},
 urldate = {2025-04-03},
 volume = {5},
 year = {2017}
}

@article{shaheen2022continual,
 author = {Shaheen, Khadija and Hanif, Muhammad Abdullah and Hasan, Osman and Shafique, Muhammad},
 journal = {Journal of Intelligent \& Robotic Systems},
 number = {1},
 pages = {9},
 publisher = {Springer},
 title = {Continual learning for real-world autonomous systems: Algorithms, challenges and frameworks},
 volume = {105},
 year = {2022}
}

@article{thompson2025social,
 author = {Thompson, Sydney and Candon, Kate and V{\'a}zquez, Marynel},
 journal = {Annual Review of Control, Robotics, and Autonomous Systems},
 publisher = {Annual Reviews},
 title = {The social context of human--robot interactions},
 volume = {9},
 year = {2025}
}

@article{tjomsland_mind_2022,
 author = {Tjomsland, Jonas and Kalkan, Sinan and Gunes, Hatice},
 doi = {10.3389/frobt.2022.669420},
 issn = {2296-9144},
 journal = {Frontiers in Robotics and AI},
 langid = {english},
 month = {March},
 publisher = {Frontiers},
 title = {Mind {{Your Manners}}! {{A Dataset}} and a {{Continual Learning Approach}} for {{Assessing Social Appropriateness}} of {{Robot Actions}}},
 urldate = {2025-01-30},
 volume = {9},
 year = {2022}
}

@inproceedings{tsoi_sean_2020,
 author = {Tsoi, Nathan and Hussein, Mohamed and Espinoza, Jeacy and Ruiz, Xavier and V{\'a}zquez, Marynel},
 booktitle = {Proceedings of the 8th {{International Conference}} on {{Human-Agent Interaction}}},
 doi = {10.1145/3406499.3418760},
 month = {November},
 pages = {281--283},
 title = {{{SEAN}}: {{Social Environment}} for {{Autonomous Navigation}}},
 urldate = {2025-04-03},
 year = {2020}
}

@article{ven_continual_2024,
  title={Continual learning and catastrophic forgetting},
  author={Van de Ven, Gido M and Soures, Nicholas and Kudithipudi, Dhireesha},
  journal={arXiv preprint arXiv:2403.05175},
  year={2024}
}

@inproceedings{walters_robotic_2007,
 author = {Walters, Michael L. and Dautenhahn, Kerstin and Woods, Sarah N. and Koay, Kheng Lee},
 booktitle = {Proceedings of the {{ACM}}/{{IEEE}} International Conference on {{Human-robot}} Interaction},
 doi = {10.1145/1228716.1228759},
 month = {March},
 pages = {317--324},
 title = {Robotic Etiquette: Results from User Studies Involving a Fetch and Carry Task},
 urldate = {2025-04-03},
 year = {2007}
}

@inproceedings{wulfmeier_incremental_2018,
 author = {Wulfmeier, Markus and Bewley, Alex and Posner, Ingmar},
 booktitle = {2018 {{IEEE International Conference}} on {{Robotics}} and {{Automation}} ({{ICRA}})},
 doi = {10.1109/ICRA.2018.8460982},
 issn = {2577-087X},
 month = {May},
 pages = {4489--4495},
 title = {Incremental {{Adversarial Domain Adaptation}} for {{Continually Changing Environments}}},
 urldate = {2025-03-17},
 year = {2018}
}

@inproceedings{zhang_adversarial_2021,
 author = {Zhang, Youshan and Davison, Brian D.},
 booktitle = {Proceedings of the International Conference on Pattern Recognition. {{Workshops}} and {{Challenges}}},
 doi = {10.1007/978-3-030-68790-8_52},
 month = {January},
 pages = {672--687},
 publisher = {Springer},
 title = {Adversarial {{Continuous Learning}} in {{Unsupervised Domain Adaptation}}},
 urldate = {2025-02-10},
 year = {2021}
}

@article{zuffer2025advancements,
 author = {Zuffer, Amara and Burke, Michael and Harandi, Mehrtash},
 journal = {arXiv preprint arXiv:2506.21899},
 title = {Advancements and Challenges in Continual Reinforcement Learning: A Comprehensive Review},
 year = {2025}
}

@article{Sandler2018MobileNetV2IR,
  title={MobileNetV2: Inverted Residuals and Linear Bottlenecks},
  author={Sandler, Mark and Howard, Andrew G. and Zhu, Menglong and Zhmoginov, Andrey and Chen, Liang-Chieh},
  journal={2018 IEEE/CVF Conference on Computer Vision and Pattern Recognition},
  year={2018},
  pages={4510-4520},
}

@inproceedings{zhang2023large,
  title={Large language models as zero-shot human models for human-robot interaction},
  author={Zhang, Bowen and Soh, Harold},
  booktitle={2023 IEEE/RSJ international conference on intelligent robots and systems (IROS)},
  pages={7961--7968},
  year={2023},
  organization={IEEE}
}

@inproceedings{10.1145/1553374.1553380,
author = {Bengio, Yoshua and Louradour, J\'{e}r\^{o}me and Collobert, Ronan and Weston, Jason},
title = {Curriculum learning},
year = {2009},
doi = {10.1145/1553374.1553380},
booktitle = {Proceedings of the 26th Annual International Conference on Machine Learning},
pages = {41–48},
}

@inproceedings{rosen2022norm,
  author={Rosen, Eric and Hsiung, Eric and Chi, Vivienne Bihe and Malle, Bertram F.},
  booktitle={IEEE RO-MAN}, 
  title={Norm Learning with Reward Models from Instructive and Evaluative Feedback}, 
  year={2022},
  pages={1634-1640},
  doi={10.1109/RO-MAN53752.2022.9900563}}

@article{checker_federated_2024,
  title = {Federated {{Learning}} of {{Socially Appropriate Agent Behaviours}} in {{Simulated Home Environments}}},
  author = {Checker, Saksham and Churamani, Nikhil and Gunes, Hatice},
  year = {2024},
  journal = {arXiv preprint arXiv.2403.07586},
}

@inproceedings{wachowiak2024llms,
  author={Wachowiak, Lennart and Coles, Andrew and Celiktutan, Oya and Canal, Gerard},
  booktitle={IEEE/RSJ IROS}, 
  title={Are Large Language Models Aligned with People’s Social Intuitions for Human–Robot Interactions?}, 
  year={2024},
  pages={2520-2527},
  doi={10.1109/IROS58592.2024.10801325}}


\appendix[Latency Analysis]
To assess the real-world deployability of our design, we measure the end-to-end inference latency, broken down per stage, over 600 randomly sampled images (100 from each domain), at batch size 1. 
Table~\ref{tab:latency_full} reports the mean, median, and tail latencies (P95/P99) for each stage of the full pipeline. The panoptic segmentation stage, done using an off-the-shelf Grounded-SAM model (GroundingDINO with SAM), dominates total latency, accounting for over 95\% of the total 7511.52 ms mean per-image latency. This cost is inherent to the segmentation backbone rather than to our architecture, as both GroundingDINO and SAM apply hard-coded input image transformations. Our own semantic decomposition step adds a comparatively small 255.02 ms overhead, and the dual-branch network itself contributes only 19.87 ms - under 0.3\% of total latency.

In Table~\ref{tab:latency_ablation} we show additional latency measurements using a single-branch ablation that bypasses panoptic segmentation and semantic decomposition. Here, total latency drops to 84.67 ms, and the forward pass itself (13.31 ms) is only 6.56 ms faster than our dual-branch design (19.87 ms). 
This shows that the dominant latency cost in our full pipeline is attributable to the segmentation model, while the overhead of our proposed dual-branch architecture with social and environmental image decomposition amounts to $<$300 ms (255.02 ms decomposition + 19.87 ms forward pass).

\begin{table}[h!]
\centering
\footnotesize
\caption{\small Per-stage inference latency (in milliseconds) of the full dual-branch pipeline. Values are mean (SD in parentheses) over $n=600$ processed images, batch size 1.}
\setlength{\tabcolsep}{3pt}
\begin{tabular}{l|cccc}
\hline
\textbf{Stage} & \textbf{Mean} & \textbf{Median} & \textbf{P95} & \textbf{P99} \\
\hline
Image Load               & 65.94 (9.69)              & 64.71   & 84.75   & 94.23 \\
Panoptic Segmentation     & 7169.36 (192.26)          & 7123.30 & 7558.30 & 7680.17 \\
Semantic Decomposition    & 255.02 (79.26)            & 240.73  & 427.27  & 525.53 \\
Input Transform           & 1.32 (0.19)               & 1.24    & 1.72    & 2.01 \\
Dual-branch Forward Pass  & 19.87 (3.40)              & 18.99   & 26.17   & 31.30 \\
\hline
\textbf{Total}            & 7511.52 (206.86) & 7479.08 & 7869.70 & 8025.08 \\
\hline
\end{tabular}
\label{tab:latency_full}
\vspace{-0.5em}
\end{table}

\begin{table}[h!]
\centering
\footnotesize
\caption{\small Per-stage inference latency (in milliseconds) of the single-image, single-branch ablation (no panoptic segmentation, no social/environment decomposition). Values are mean (SD in parentheses) over $n=600$ processed images, batch size 1.}
\setlength{\tabcolsep}{4.5pt}
\begin{tabular}{l|cccc}
\hline
\textbf{Stage} & \textbf{Mean} & \textbf{Median} & \textbf{P95} & \textbf{P99} \\
\hline
Image Load                & 70.30 (8.85)          & 69.35 & 85.66 & 96.88 \\
Segmentation (bypassed)   & 0.014 (0.004)         & 0.012 & 0.020 & 0.024 \\
Decomposition (bypassed)  & 0.013 (0.008)         & 0.011 & 0.020 & 0.029 \\
Input Transform           & 1.03 (0.18)           & 1.01  & 1.31  & 1.41 \\
Single-branch Forward Pass & 13.31 (4.33)          & 12.66 & 21.48 & 27.94 \\
\hline
\textbf{Total}             & 84.67 (10.15) & 83.47 & 101.62 & 120.19 \\
\hline
\end{tabular}
\label{tab:latency_ablation}
\vspace{-0.5em}
\end{table}

\end{document}